\documentclass[10pt,twocolumn,letterpaper]{article}

\usepackage[pagenumbers]{cvpr} 

\input{preamble}

\definecolor{cvprblue}{rgb}{0.21,0.49,0.74}
\usepackage[pagebackref,breaklinks,colorlinks,citecolor=cvprblue]{hyperref}

\def\paperID{X} 
\def\confName{CVPR}
\def\confYear{2024}

\title{RoMa: Robust Dense Feature Matching}

\author{Johan Edstedt$^1$
\quad
 Qiyu Sun$^2$
\quad
Georg Bökman$^3$
\quad
 Mårten Wadenbäck$^1$
\quad
 Michael Felsberg$^1$
 \\
{\normalsize $^1$Linköping University,
$^2$East China University of Science and Technology,
$^3$Chalmers University of Technology}
}

\begin{document}
\twocolumn[{%
\renewcommand\twocolumn[1][]{#1}%
\maketitle
\includegraphics[width=.24\textwidth]{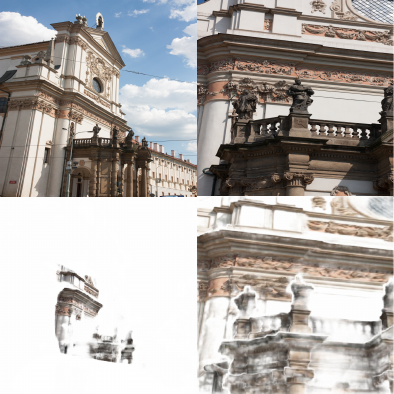}
\hfill
\includegraphics[width=.24\textwidth]{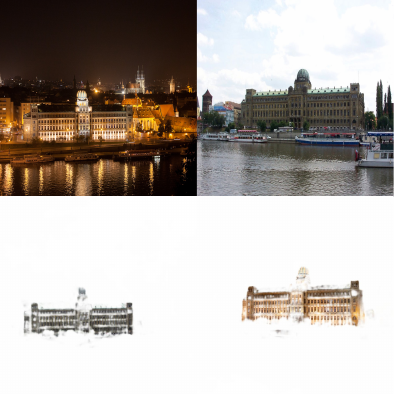}
\hfill
\includegraphics[width=.24\textwidth]{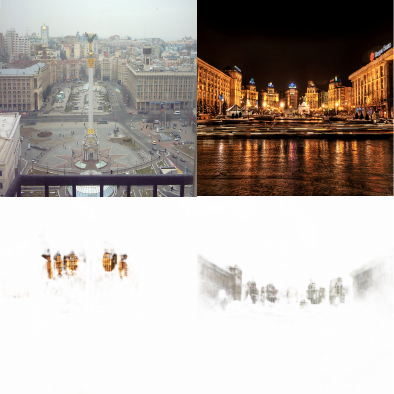}
\hfill
\includegraphics[width=.24\textwidth]{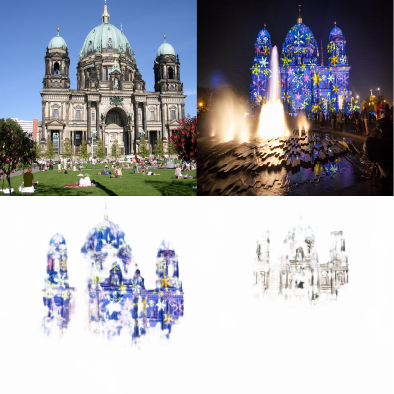}
\hfill
\captionof{figure}{\textbf{\ours~is robust, \ie, able to match under extreme changes.} We propose \ours, a model for dense feature matching that is robust to a wide variety of challenging real-world changes in scale, illumination, viewpoint, and texture. We show correspondences estimated by \ours~on the extremely challenging benchmark WxBS~\citep{Mishkin2015WXBS}, where most previous methods fail, and on which we set a new state-of-the-art with an improvement of \textbf{36\% mAA}. The estimated correspondences are visualized by grid sampling coordinates bilinearly from the other image, using the estimated warp, and multiplying with the estimated confidence.\vspace{2em}}
\label{fig:teaser}}]


\begin{abstract}
Feature matching is an important computer vision task that involves estimating correspondences between two images of a 3D scene, and dense methods estimate all such correspondences. The aim is to learn a robust model, \emph{\ie}, a model able to match under challenging real-world changes. 
In this work, we propose such a model, leveraging frozen pretrained features from the foundation model DINOv2. Although these features are significantly more robust than local features trained from scratch, they are inherently coarse. We therefore combine them with specialized ConvNet fine features, creating a precisely localizable feature pyramid. To further improve robustness, we propose a tailored transformer match decoder that predicts anchor probabilities, which enables it to express multimodality. Finally, we propose an improved loss formulation through regression-by-classification with subsequent robust regression. We conduct a comprehensive set of experiments that show that our method, \ours, achieves significant gains, setting a new state-of-the-art. In particular, we achieve a 36\% improvement on the extremely challenging WxBS benchmark. Code is provided at \href{https://github.com/Parskatt/RoMa}{github.com/Parskatt/RoMa}.
\end{abstract}


\section{Introduction}
\label{sec:intro}
\begin{figure*}
    \centering
\includegraphics[width=\linewidth]{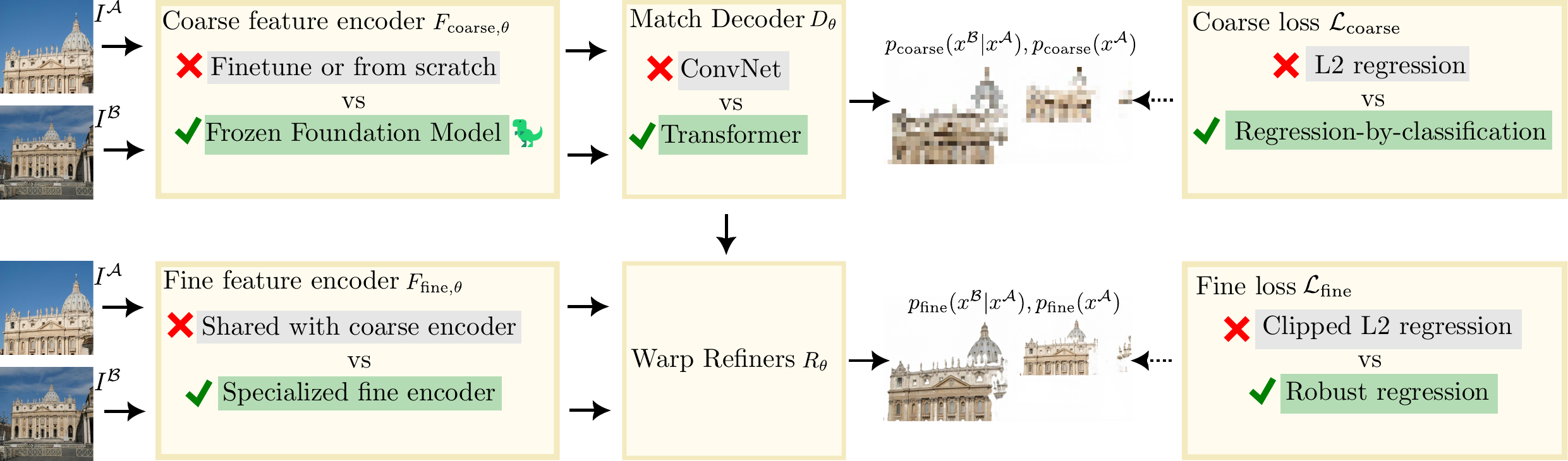}
    \caption{\textbf{Illustration of our robust approach \ours.} Our contributions are shown with green highlighting and a checkmark, while previous approaches are indicated with gray highlights and a cross. Our first contribution is using a frozen foundation model for coarse features, compared to fine-tuning or training from scratch. DINOv2 lacks fine features, which are needed for accurate correspondences. To tackle this, we combine the DINOv2 coarse features with specialized fine features from a ConvNet, see Section~\ref{sec:features}. Second, we propose an improved coarse match decoder $D_{\theta}$, which typically is a ConvNet, with a coordinate agnostic Transformer decoder that predicts anchor probabilities instead of directly regressing coordinates, see Section~\ref{sec:transformer-decoder}. Third, we revisit the loss functions used for dense feature matching. We argue from a theoretical model that the global matching stage needs to model multimodal distributions, and hence use a regression-by-classification loss instead of an L2 loss. For the refinement, we in contrast use a robust regression loss, as the matching distribution is locally unimodal. These losses are further discussed in Section~\ref{sec:loss}. The impact of our contributions is ablated in our extensive ablation study in Table~\ref{tab:ablation}.}
    \label{fig:method}
\end{figure*}
Feature matching is the computer vision task of from two images estimating pixel pairs that correspond to the same 3D point. It is crucial for downstream tasks such as 3D reconstruction~\citep{schonberger2016structure} and visual localization~\citep{sarlin2019coarse}. 
Dense feature matching methods~\citep{truong2020gocor, truong2023pdc, edstedt2023dkm, ni2023pats} aim to find all matching pixel-pairs between the images. These dense methods employ a coarse-to-fine approach, whereby matches are first predicted at a coarse level and successively refined at finer resolutions. 
Previous methods commonly learn coarse features using 3D supervision~\citep{sarlin2020superglue,sun2021loftr,truong2023pdc,edstedt2023dkm}. While this allows for specialized coarse features, it comes with downsides. In particular, since collecting real-world 3D datasets is expensive, the amount of available data is limited, which means models risk overfitting to the training set. This in turn limits the models robustness to scenes that differ significantly from what has been seen during training.
A well-known approach to limit overfitting is to freeze the backbone used~\citep{tian2020prior,vasconcelos2022proper,lin2022could}. However, using frozen backbones pretrained on ImageNet classification, the out-of-the-box performance is insufficient for feature matching (see experiments in Table~\ref{tab:linprobe}). A recent promising direction for frozen pretrained features is large-scale self-supervised pretraining using Masked image Modeling (MIM)~\citep{he2022mae, wei2022masked, zhou2022image, oquab2023dinov2}. 
The methods, including DINOv2~\citep{xie2023revealing}, retain local information better than classification pretraining~\citep{xie2023revealing} and have been shown to generate features that generalize well to dense vision tasks. However, the application of DINOv2 in dense feature matching is still complicated due to the lack of fine features, which are needed for refinement.

We overcome this issue by leveraging a frozen DINOv2 encoder for coarse features, while using a proposed specialized ConvNet encoder for the fine features. This has the benefit of incorporating the excellent general features from DINOv2, while simultaneuously having highly precise fine features. We find that features specialized for only coarse matching or refinement significantly outperform features trained for both tasks jointly. These contributions are presented in more detail in Section~\ref{sec:features}. We additionally propose a Transformer match decoder that while also increasing performance for the baseline, particularly improves performance when used to predict anchor probabilities instead of regressing coordinates in conjunction with the DINOv2 coarse encoder. This contribution is elaborated further in Section~\ref{sec:transformer-decoder}.

Lastly, we investigate how to best train dense feature matchers. Recent SotA dense methods such as DKM~\citep{edstedt2023dkm} use a non-robust regression loss for the coarse matching as well as for the refinement. We argue that this is not optimal as the matching distribution at the coarse stage is often multimodal, while the conditional refinement is more likely to be unimodal. Hence requiring different approaches to training. We motivate this from a theoretical framework in Section~\ref{sec:loss}. Our framework motivates a division of the coarse and fine losses into seperate paradigms, regression-by-classification for the global matches using coarse features, and robust regression for the refinement using fine features. 

Our full approach, which we call \ours, is robust to extremely challenging real-world cases, as we demonstrate in Figure~\ref{fig:teaser}. We illustrate our approach schematically in Figure~\ref{fig:method}. 
In summary, our contributions are as follows:
\begin{enumerate}[topsep=0.5ex,itemsep=0ex,label=(\alph*)]
    \item We integrate frozen features from the foundation model DINOv2~\citep{oquab2023dinov2} for dense feature matching. We combine the coarse features from DINOv2 with specialized fine features from a ConvNet to produce a precisely localizable yet robust feature pyramid. See Section~\ref{sec:features}.
    \item We propose a Transformer-based match decoder, which predicts anchor probabilities instead of coordinates. See Section~\ref{sec:transformer-decoder}.
    \item We improve the loss formulation. In particular, we use a regression-by-classification loss for coarse global matches, while we use robust regression loss for the refinement stage, both of which we motivate from a theoretical analysis. See Section~\ref{sec:loss}.
    \item We conduct an extensive ablation study over our contributions, and SotA experiments on a set of diverse and competitive benchmarks, and find that \ours~sets a new state-of-the-art. In particular, achieving a gain of 36\% on the difficult WxBS benchmark.  See Section~\ref{sec:experiments}.
\end{enumerate}
\section{Related Work}
\label{sec:formatting}
\subsection{Sparse \texorpdfstring{$\to$}{->} Detector Free \texorpdfstring{$\to$}{->} Dense Matching}
Feature matching has traditionally been approached by keypoint detection and description followed by matching the descriptions~\citep{lowe2004distinctive,bay2006surf,detone2018superpoint,revaud2019r2d2,sarlin2020superglue,tyszkiewicz2020disk}.
Recently, the detector-free approach~\citep{sun2021loftr,bokman2022case,tang2022quadtree,chen2022aspanformer} replaces the keypoint detection with dense matching on a coarse scale, followed by mutual nearest neighbors extraction, which is followed by refinement. 
The dense approach~\citep{melekhov2019dgc,truong2020glu,truong2021learning,edstedt2023dkm,zhu2023pmatch,ni2023pats} instead estimates a dense warp, aiming to estimate every matchable pixel pair.

\subsection{Self-Supervised Vision Models}
Inspired by language Transformers~\citep{devlin2019bert}  foundation models~\citep{bommasani2021opportunities} pre-trained on large quantities of data have recently demonstrated significant potential in learning all-purpose features for various visual models via self-supervised learning.
Caron et al.~\citep{caron2021emerging} observe that
self-supervised ViT features capture more distinct information than supervised models do, which is demonstrated through label-free self-distillation.
iBOT~\citep{zhou2022image} explores MIM within a self-distillation framework to develop a semantically rich visual tokenizer, yielding robust features effective in various dense downstream tasks.
DINOv2~\citep{oquab2023dinov2} reveals that self-supervised methods can produce all-purpose visual features that work across various image distributions and tasks after being trained on sufficient datasets without finetuning.


\subsection{Robust Loss Formulations}
\parsection{Robust Regression Losses:}
Robust loss functions provide a continuous transition between an inlier distribution (typically highly concentrated), and an outlier distribution (wide and flat). 
Robust losses have, \eg, been used as regularizers for optical flow~\citep{black1996robust,black1996unification}, robust smoothing~\citep{felsberg2006channel}, and as loss functions~\citep{barron2019general,liu2022camliflow}. 

\parsection{Regression by Classification:} 
Regression by classification~\citep{weiss1993rbc,weiss1995rbc,luis1996rbc} involves casting regression problems as classification by, \eg, binning. This is particularly useful for regression problems with sharp borders in motion, such as stereo disparity~\citep{garg2020wasserstein,hager2021predicting}. \citet{germain2021neural} use a regression-by-classification loss for absolute pose regression.

\parsection{Classification then Regression:}
\citet{li2020hierarchical}, and \citet{budvytis2019largescale} proposed hierarchical classification-regression frameworks for visual localization. \citet{sun2021loftr} optimize the model log-likelihood of mutual nearest neighbors, followed by L2 regression-based refinement for feature matching.
\newpage
\section{Method}
\label{sec:method}
In this section, we detail our method. We begin with preliminaries and notation for dense feature matching in Section~\ref{sec:preliminaries}. We then discuss our incorporation of DINOv2~\citep{oquab2023dinov2} as a coarse encoder, and specialized fine features in Section~\ref{sec:features}. We present our proposed Transformer match decoder in Section~\ref{sec:transformer-decoder}. Finally, our proposed loss formulation in Section~\ref{sec:loss}. A summary and visualization of our full approach is provided in Figure~\ref{fig:method}. Further details on the exact architecture are given in the supplementary.
\subsection{Preliminaries on Dense Feature Matching}
\label{sec:preliminaries}
 Dense feature matching is, given two images $I^\A , I^\B$, to estimate a dense warp $W^{\A\to\B}$ (mapping coordinates $x^\A$ from $I^\A$ to $x^\B$ in $I^\B$), and a matchability score $p(x^{\A})$\footnote{This is denoted as $p^{\A\to\B}$ by~\citet{edstedt2023dkm}. We omit the $\B$ to avoid confusion with the conditional.} for each pixel. From a probabilistic perspective, $p(W^{\A\to\B}) = p(x^{\B}|x^{\A})$ is the conditional matching distribution. Multiplying $p(x^{\B}|x^{\A})p(x^{\A})$ yields the joint distribution. We denote the model distribution as $p_{\theta}(x^{\A},x^{\B}) = p_{\theta}(x^{\B}|x^{\A})p_{\theta}(x^{\A})$. When working with warps, \ie, where $p_{\theta}(x^{\B}|x^{\A})$ has been converted to a deterministic mapping, we denote the model warp as $\hat{W}^{\A\to\B}$. Viewing the predictive distribution as a warp is natural in high resolution, as it can then be seen as a deterministic mapping. However, due to multimodality, it is more natural to view it in the probabilistic sense at coarse scales.
 
 The end goal is to obtain a good estimate over
 correspondences of coordinates $x^\A$ in image $I^\A$ and coordinates $x^\B$ in image $I^\B$. 
 For dense feature matchers, estimation of these correspondences is typically done by a one-shot coarse \emph{global matching} stage (using coarse features) followed by subsequent \emph{refinement} of the estimated warp and confidence (using fine features). 
 
 We use the recent SotA dense feature matching model DKM~\citep{edstedt2023dkm} as our baseline. For consistency, we adapt the terminology used there. We denote the coarse features used to estimate the initial warp, and the fine features used to refine the warp by
\begin{equation}
    \{\varphi^{\A}_{\text{coarse}},\varphi^{\A}_{\text{fine}}\} = F_{\theta}(I^{\A}),
        \{\varphi^{\B}_{\text{coarse}},\varphi^{\B}_{\text{fine}}\} = F_{\theta}(I^{\B}),
\end{equation}
 where $F_{\theta}$ is a neural network feature encoder.  We will leverage DINOv2 for extraction of $\varphi^{\A}_{\text{coarse}}, \varphi^{\B}_{\text{coarse}}$, however, DINOv2 features are not precisely localizable, which we tackle by combining the coarse features with precise local features from a specialized ConvNet backbone. See Section~\ref{sec:features} for details.

The coarse features are matched with global matcher $G_{\theta}$ consisting of a match encoder $E_{\theta}$ and match decoder $D_{\theta}$, 
\begin{equation}
\left\{
    \begin{aligned}
    \big(\hat{\text{W}}^{\A\to\B}_{\text{coarse}},\enspace p^{\A}_{\theta,\text{coarse}}\big) &= G_{\theta}(\varphi^{\A}_{\text{coarse}},\varphi^{\B}_{\text{coarse}}),\enspace\\
    G_{\theta}({\varphi}^{\A}_{\text{coarse}},{\varphi}^{\B}_{\text{coarse}}) &= D_{\theta}\big(E_{\theta}({\varphi}^{\A}_{\text{coarse}},{\varphi}^{\B}_{\text{coarse}})\big).
\end{aligned}
\right.
\end{equation}
We use a Gaussian Process~\citep{rasmussen2003gaussian} as the match encoder $E_{\theta}$ as in previous work~\citep{edstedt2023dkm}. However, while our baseline uses a ConvNet to decode the matches, we propose a Transformer match decoder $D_{\theta}$ that predicts anchor probabilities instead of directly regressing the warp. This match decoder is particularly beneficial in our final approach (see Table~\ref{tab:ablation}). We describe our proposed match decoder in Section~\ref{sec:transformer-decoder}.
The refinement of the coarse warp $\hat{\text{W}}^{\A\to\B}_{\text{coarse}}$ is done by the refiners $R_{\theta}$,
\begin{equation}
    \big(\hat{\text{W}}^{\A\to\B},\, p_{\theta}^{\A}\big) = R_{\theta}\big(\varphi_{\text{fine}}^{\A},\varphi_{\text{fine}}^{\B},\hat{\text{W}}^{\A\to\B}_{\text{coarse}},p^{\A}_{\theta,\text{coarse}}\big).
\end{equation}
As in previous work, the refiner is composed of a sequence of ConvNets (using strides $\{1,2,4,8\}$) and can be decomposed recursively as
\begin{equation}
    \big(\hat{W}^{\A\to\B}_i,\;p_{i,\theta}^{\A}\big)   = R_{\theta,i}(\varphi_i^{\A},\varphi_i^{\B},\hat{W}^{\A\to\B}_{i+1},p_{\theta,i+1}^{\A}),
\end{equation}
where the stride is $2^i$. 
The refiners predict a residual offset for the estimated warp, and a logit offset for the certainty. As in the baseline they are conditioned on the outputs of the previous refiner by using the previously estimated warp to a) stack feature maps from the images, and b) construct a local correlation volume around the previous target.

The process is repeated until reaching full resolution. We use the same architecture as in the baseline. Following DKM, we detach the gradients between the refiners and 
upsample the warp bilinearly to match the resolution of the finer stride.

\parsection{Probabilistic Notation:}
When later defining our loss functions, it will be convenient to refer to the outputs of the different modules in a probabilistic notation. We therefore introduce this notation here first for clarity.

We denote the probability distribution modeled by the global matcher as
\begin{equation}
    p_{\theta}(x_{\text{coarse}}^{\A},x_{\text{coarse}}^{\B}) = G_{\theta}(\varphi_{\text{coarse}}^{\A},\varphi_{\text{coarse}}^{\B}).
\end{equation}
Here we have dropped the explicit dependency on the features and the previous estimate of the marginal for notational brevity. Note that the output of the global matcher will sometimes be considered as a discretized distribution using anchors, or as a decoded warp. We do not use separate notation for these two different cases to keep the notation uncluttered.

We denote the probability distribution modeled by a refiner at scale $s=c2^i$ as
\begin{equation}
    p_{\theta}(x_i^{\A},x_i^{\B}|\hat{W}^{\A\to\B}_{i+1}) = R_{\theta,i}(\varphi_i^{\A},\varphi_i^{\B},\hat{W}^{\A\to\B}_{i+1},p_{\theta,i+1}^{\A}),
\end{equation}
The basecase $\hat{W}^{\A\to\B}_{\text{coarse}}$ is computed by decoding $p_{\theta}(x_{\text{coarse}}^{\B}|x_{\text{coarse}}^{\A})$.
As for the global matcher we drop the explicit dependency on the features.

\subsection{Robust and Localizable Features}
\label{sec:features}

We first investigate the robustness of DINOv2 to viewpoint and illumination changes compared to VGG19 and ResNet50 on the MegaDepth~\citep{li2018megadepth} dataset. To decouple the backbone from the matching model we train a single linear layer on top of the frozen model followed by a kernel nearest neighbour matcher for each method. We measure the performance both in average end-point-error (EPE) on a standardized resolution of 448$\times$448, and by what we call the Robustness $\%$ which we define as the percentage of matches with an error lower than $32$ pixels. We refer to this as robustness, as, while these matches are not necessarily accurate, it is typically sufficient for the refinement stage to produce a correct adjustment.

We present results in Table~\ref{tab:linprobe}. We find that DINOv2 features are significantly more robust to changes in viewpoint than both ResNet and VGG19. Interestingly, we find that the VGG19 features are worse than the ResNet features for coarse matching, despite VGG feature being widely used as local features~\citep{dusmanu2019d2,sarlin2021back,truong2023pdc}. Further details of this experiment are provided in the supplementary material.
\begin{table}
    \centering
    \small
    \caption{\textbf{Evaluation of frozen features on MegaDepth.} We compare the VGG19 and ResNet50 backbones commonly used in feature matching with the generalist features of DINOv2.}
    \begin{tabular}{l ll}
        \toprule
        Method & EPE $\downarrow$ & Robustness \% $\uparrow$\\
        \midrule
           VGG19 & 87.6 & 43.2\\
          RN50 & 60.2 & 57.5\\
          DINOv2 & \textbf{27.1} & \textbf{85.6}\\
          \bottomrule
    \end{tabular}
    \label{tab:linprobe}
\end{table}

In DKM~\citep{edstedt2023dkm}, the feature encoder $F_{\theta}$ is assumed to consist of a single network producing a feature pyramid of coarse and fine features used for global matching and refinement respectively. This is problematic when using DINOv2 features as only features of stride 14 exist. We therefore decouple $F_{\theta}$ into $\{F_{\text{coarse},\theta}, F_{\text{fine},\theta}\}$ and set $F_{\text{coarse},\theta} = \text{DINOv2}$. The coarse features are extracted as
\begin{equation}
    \varphi^{\A}_{\text{coarse}} = F_{\text{coarse},\theta}(I^{\A}),
        \varphi^{\B}_{\text{coarse}} = F_{\text{coarse},\theta}(I^{\B}).
\end{equation}
We keep the DINOv2 encoder frozen throughout training. This has two benefits. The main benefit is that keeping the representations fixed reduces overfitting to the training set, enabling \ours~to be more robust. It is also additionally significantly cheaper computationally and requires less memory.
However, DINOv2 cannot provide fine features. Hence a choice of $F_{\text{fine},\theta}$ is needed. While the same encoder for fine features as in DKM could be chosen, \ie, a ResNet50 (RN50)~\citep{he2016deep}, it turns out that this is not optimal. 

We begin by investigating what happens by simply decoupling the coarse and fine feature encoder, \ie, not sharing weights between the coarse and fine encoder (even when using the same network). We find that, as supported by Setup~\hyperref[case:II]{II} in Table~\ref{tab:ablation}, this significantly increases performance. This is due to the feature extractor being able to specialize in the respective tasks, and hence call this \emph{specialization}.

This raises a question, VGG19 features, while less suited for coarse matching (see Table~\ref{tab:linprobe}), could be better suited for fine localized features. We investigate this by setting $F_{\text{fine},\theta} = \text{VGG19}$ in Setup~\hyperref[case:III]{III} in Table~\ref{tab:ablation}. Interestingly, even though VGG19 coarse features are significantly worse than RN50, we find that they significantly outperform the RN50 features when leveraged as fine features. 
Our finding indicates that there is an inherent tension between fine localizability and coarse robustness. We thus use VGG19 fine features in our full approach. 
\subsection{Transformer Match Decoder $D_{\theta}$}
\label{sec:transformer-decoder}

\parsection{Regression-by-Classification:}
We propose to use the \emph{regression by classification} formulation for the match decoder, whereby we discretize the output space. We choose the following formulation,
\begin{equation}
\label{eq:coarse-prob}
    p_{\text{coarse},\theta}(x^{\B}|x^{\A}) = \sum_{k=1}^K \pi_k(x^{\A}) \mathcal{B}_{m_k},
\end{equation}
where $K$ is the quantization level, $\pi_k$ are the probabilities for each component, $\mathcal{B}$ is some 2D base distribution, and $\{m_k\}_1^K$ are anchor coordinates. 
In practice, we used $K = 64\times 64$ classification anchors positioned uniformly as a tight cover of the image grid, and $\mathcal{B} = \mathcal{U}$, \ie, a uniform distribution\footnote{This ensures that there is no overlap between anchors and no holes in the cover.}. We denote the probability of an anchor as $\pi_k$ and its associated coordinate on the grid as $m_k$.

For refinement, the conditional is converted to a deterministic warp per pixel. We decode the warp by argmax over the classification anchors, $k^*(x) = \argmax_k \pi_k(x)$, followed by a local adjustment which can be seen as a local softargmax. Mathematically,
\begin{align}
\label{eq:to-warp}
    \text{ToWarp}(p_{\text{coarse},\theta}(x^{\B}_{\text{coarse}}|x^{\A}_{\text{coarse}})) =\nonumber\\
    \frac{\sum_{i\in N_4(k^*(x^{\A}_{\text{coarse}}))}\pi_{i} m_{i}}{\sum_{i\in N_4(k^*(x^{\A}_{\text{coarse}}))}\pi_{i}} =\hat{W}^{\A\to\B}_{\text{coarse}},
\end{align}
where $N_4(k^*)$ denotes the set of $k^*$ and the four closest anchors on the left, right, top, and bottom.
We conduct an ablation on the Transformer match decoder in Table~\ref{tab:ablation}, and find that it particularly improves results in our full approach, using the loss formulation we propose in Section~\ref{sec:loss}.

\parsection{Decoder Architecture:}
In early experiments, we found that ConvNet coarse match decoders overfit to the training resolution. Additionally, they tend to be over-reliant on locality. While locality is a powerful cue for refinement, it leads to oversmoothing for the coarse warp. To address this, we propose a transformer decoder without using position encodings. By restricting the model to only propagate by feature similarity, we found that the model became significantly more robust. 

The proposed Transformer matcher decoder consists of 5 ViT blocks, with 8 heads, hidden size D 1024, and MLP size 4096.
The input is the concatenation of projected DINOv2~\citep{oquab2023dinov2} features of dimension 512, and the 512-dimensional output of the GP module, which corresponds to the match encoder $E_{\theta}$ proposed in DKM~\citep{edstedt2023dkm}. The output is a vector of $B\times H\times W\times (K + 1)$ where $K$ is the number of classification anchors\footnote{When used for regression, $K$ is set to $K=2$, and the decoding to a warp is the identity function.} (parameterizing the conditional distribution $p(x^{\B}|x^{\A})$), and the extra 1 is the matchability score $p^{\A}(x^{\A})$ .

\subsection{Robust Loss Formulation}
\label{sec:loss}

\parsection{Intuition:}
The conditional match distribution at coarse scales is more likely to exhibit multimodality than during refinement, which is conditional on the previous warp. This means that the coarse matcher needs to model multimodal distributions, which motivates our regression-by-classification approach. In contrast, the refinement of the warp needs only to represent unimodal distributions, which motivates our robust regression loss.

\parsection{Theoretical Model:}
\begin{figure}
    \centering
    \includegraphics[width=\linewidth]{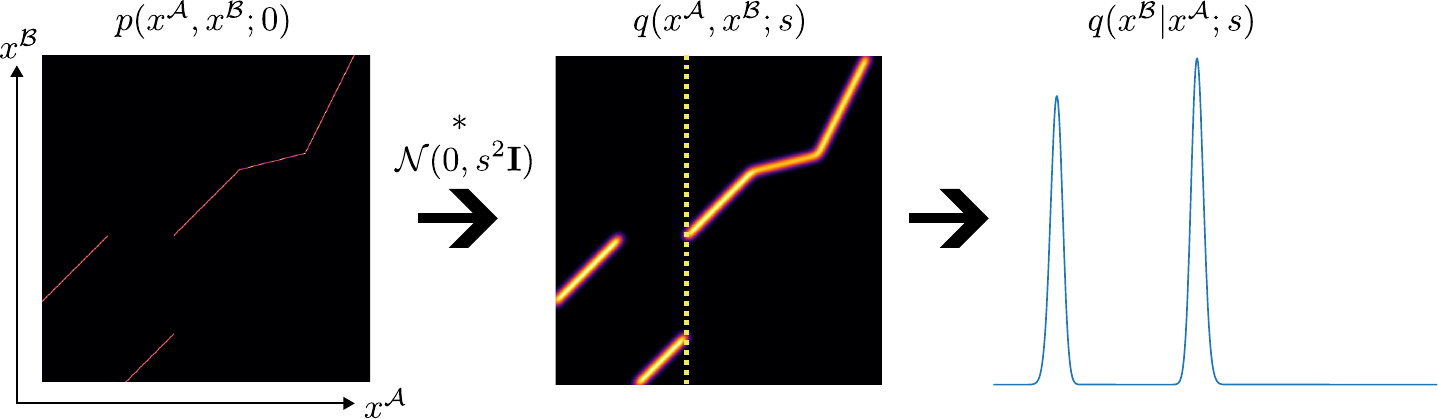}
    \caption{\textbf{Illustration of localizability of matches.} At infinite resolution the match distribution can be seen as a 2D surface (illustrated as 1D lines in the figure), however at a coarser scale $s$ this distribution becomes blurred due to motion boundaries. This means it is necessary to both use a model and an objective function capable of representing multimodal distributions.}
    \label{fig:scalespace}
\end{figure}
We model the matchability at scale $s$ as 
 \begin{equation}
     q(x^{\A},x^{\B}; s) = \mathcal{N}\big(0,s^2\mathbf{I}\big) \ast p(x^{\A},x^{\B};0).
 \end{equation}
 Here $p(x^{\A},x^{\B};0)$ corresponds to the exact mapping at infinite resolution.
This can be interpreted as a diffusion in the localization of the matches over scales. When multiple objects in a scene are projected into images, so-called motion boundaries arise. These are discontinuities in the matches which we illustrate in Figure~\ref{fig:scalespace}. The diffusion near these motion boundaries causes the conditional distribution to become multimodal, explaining the need for multimodality in the coarse global matching.
 Given an initial choice of $(x^{\A}, x^{\B})$, as in the refinement, the conditional distribution is unimodal locally. However, if this initial choice is far outside the support of the distribution, using a non-robust loss function is problematic. It is therefore motivated to use a robust regression loss for this stage.

\parsection{Loss formulation:}
Motivated by intuition and the theoretical model we now propose our loss formulation from a probabilistic perspective, aiming to minimize the Kullback--Leibler divergence between the estimated match distribution at each scale, and the theoretical model distribution at that scale. We begin by formulating the coarse loss. With non-overlapping bins as defined in Section~\ref{sec:transformer-decoder} the Kullback--Leibler divergence (where terms that are constant w.r.t.~ $\theta$ are ignored) is
\begin{align}
D_{\rm KL}(q(x^{\B},x^{\A};s)|| p_{\text{coarse},\theta}(x^{\B},x^{\A})) =\\
    \mathbb{E}_{x^{\A}, x^{\B}\sim q} \big[-\log p_{\text{coarse},\theta}(x^{\B}|x^{\A}) p_{\text{coarse},\theta}(x^{\A})\big] =\\ 
    -\int_{x^{\A}, x^{\B}}\log \pi_{k^{{\dag}}}(x^{\A}) + \log p_{\text{coarse},\theta}(x^{\A})dq,
\end{align}
for $k^{\dag}(x) = {\rm argmin}_k \lVert m_k-x\rVert$ the index of the closest anchor to $x$. 
Following DKM~\citep{edstedt2023dkm} we add a hyperparameter $\lambda$ that controls the weighting of the marginal compared to that of the conditional as
\begin{equation}
        -\int_{x^{\A}, x^{\B}}\log \pi_{k^{\dag}}(x^{\A}) + \lambda \log p_{\text{coarse},\theta}(x^{\A})dq.
\end{equation}
\begin{figure}
    \centering
    \includegraphics[width=.8\linewidth]{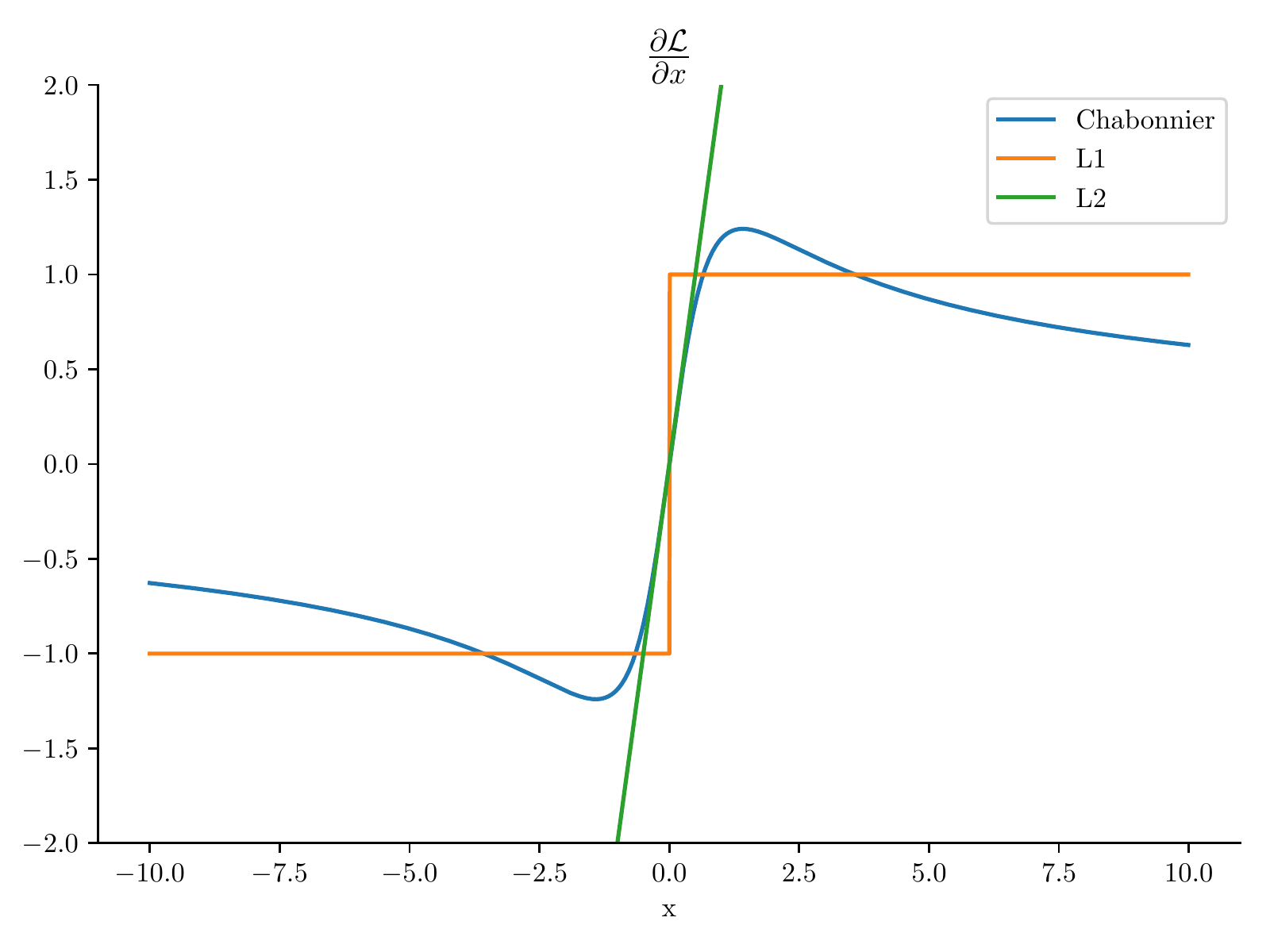}
    \caption{\textbf{Comparison of loss gradients.} We use the generalized Charbonnier~\citep{barron2019general} loss for refinement, which locally matches L2 gradients, but globally decays with $|x|^{-1/2}$ toward zero.}
    \label{fig:robust-loss}
\end{figure}

In practice, we approximate $q$ with a discrete set of known correspondences $\{x^{\A}, x^{\B}\}$. Furthermore, to be consistent with previous works~\citep{truong2023pdc, edstedt2023dkm} we use a binary cross-entropy loss on $p_{\text{coarse},\theta}(x^{\A})$. 
We call this loss $\mathcal{L}_{\text{coarse}}$. We next discuss the fine loss $\mathcal{L}_{\text{fine}}$.

We model the output of the refinement at scale $i$ as a generalized Charbonnier~\citep{barron2019general} (with $\alpha = 0.5$) distribution, for which the refiners estimate the mean $\mu$. The generalized Charbonnier distribution behaves locally like a Normal distribution, but has a flatter tail. When used as a loss, the gradients behave locally like L2, but decay towards 0, see Figure~\ref{fig:robust-loss}. Its logarithm, (ignoring terms that do not contribute to the gradient, and up-to-scale) reads
\begin{align}
    \log p_{\theta}(x_i^{\B}|x_i^{\A},\hat{W}^{\A\to\B}_{i+1}) =\\ -(||\mu_{\theta}(x_i^{\A},\hat{W}^{\A\to\B}_{i+1})-x_i^{\B}||^2 + s)^{1/4},
\end{align}
where $\mu_{\theta}(x_{i}^{\A},\hat{W}^{\A\to\B}_{i+1})$ is the estimated mean of the distribution, and $s=2^ic$. In practice, we choose $c=0.03$.
The Kullback--Leibler divergence for each fine scale $i\in \{0, 1, 2, 3\}$ (where terms that are constant with respect to $\theta$ are ignored) reads
\begin{align}
    D_{\rm KL}(q(x_i^{\B},x_i^{\A};s=2^i c)|| p_{i,\theta}(x_i^{\B},x_i^{\A}|\hat{W}^{\A\to\B}_{i+1})) =\\
    \mathbb{E}_{x_i^{\A}, x_i^{\B}\sim q}\big[-(||\mu_{\theta}(x_i^{\A},\hat{W}^{\A\to\B}_{i+1})-x_i^{\B}||^2 + s)^{1/4}\big] +\nonumber\\
    \mathbb{E}_{x_i^{\A}, x_i^{\B}\sim q}\big[-\log p_{i,\theta}(x_i^{\A}|\hat{W}^{\A\to\B}_{i+1})\big].
\end{align}
In practice, we approximate $q$ with a discrete set of known correspondences $\{x^{\A}, x^{\B}\}$, and use a binary cross-entropy loss on $p_{\text{coarse},\theta}(x_i^{\A}|\hat{W}^{\A\to\B}_{i+1})$. We sum over all fine scales to get the loss $\mathcal{L}_{\text{fine}}$.

Our combined loss yields:
\begin{equation}
    \mathcal{L} = \mathcal{L}_{\text{coarse}} + \mathcal{L}_{\text{fine}}.
\end{equation}

Note that we do not need to tune any scaling between these losses as the coarse matching and fine stages are decoupled as gradients are cut in the matching, and encoders are not shared.

\begin{table}
    \small
    \centering
    \caption{\textbf{Ablation study.} We systematically investigate the impact of our contributions, see Section~\ref{sec:ablation} for detailed analysis. Measured in 100-percentage correct keypoints (PCK) (lower is better).}
    \begin{tabular}{l llll}
    \toprule
         Setup $\downarrow$\quad\quad\quad 100-PCK$@$ $\rightarrow$       & 1px  & 3px  & 5px   \\
         \midrule
    \rowcolor{gray!25}\label{case:I}I (Baseline): DKM~\citep{edstedt2023dkm}  &	17.0 & 7.3 & 5.8 \\
    \label{case:II}II: I, $F_{\text{coarse},\theta} = \text{RN50}$, $F_{\text{fine},\theta} = \text{RN50}$ &	16.0 & 6.1 & 4.5 \\
    \label{case:III}III: II, $F_{\text{fine}, \theta} = \text{VGG19}$ & 14.5 & 5.4 & 4.5 \\
    \label{case:IV}IV: III,  $D_{\theta} = \text{Transformer}$& 14.4 & 5.4 & 4.1 \\
    \label{case:V}V: IV, $F_{\text{coarse},\theta} = \text{DINOv2}$ & 14.3 & 4.6 & 3.2 \\
    \label{case:VI}VI: V, $\mathcal{L}_{\text{coarse}} = \text{reg.-by-class.}$ &	13.6 & 4.1 & 2.8\\
   \rowcolor{green!25} \label{case:VII}\textbf{VII (Ours):}  VI, $\mathcal{L}_{\text{refine}} = \text{robust}$   & \textbf{13.1}  & \textbf{4.0} & \textbf{2.7}\\
    \label{case:VIII}VIII: VII, $D_{\theta} = \text{ConvNet}$  &14.0 & 4.9 & 3.5\\ 
         \bottomrule
    \end{tabular}

    \label{tab:ablation}
\end{table}

\section{Experiments}
\label{sec:experiments}
\subsection{Ablation Study}
\label{sec:ablation}
Here we investigate the impact of our contributions. We conduct all our ablations on a validation test that we create. The validation set is made from random pairs from the MegaDepth scenes $[0015,0022]$ with overlap $>0$. To measure the performance we measure the percentage of estimated matches that have an end-point-error (EPE) under a certain pixel threshold over all ground-truth correspondences, which we call percent correct keypoints (PCK) using the notation of previous work~\citep{truong2023pdc,edstedt2023dkm}. 

Setup \hyperref[case:I]{I} consists of the same components as in DKM~\citep{edstedt2023dkm}, retrained by us. In Setup \hyperref[case:II]{II} we do not share weights between the fine and coarse features, which improves performance due to specialization of the features. In Setup \hyperref[case:III]{III} we replace the RN50 fine features with a VGG19, which further improves performance. This is intriguing, as VGG19 features are worse performing when used as coarse features as we show in Table~\ref{tab:linprobe}. We then add the proposed Transformer match decoder in Setup \hyperref[case:IV]{IV}, however using the baseline regression approach. Further, we incorporate the DINOv2 coarse features in Setup \hyperref[case:V]{V}, this gives a significant improvement, owing to their significant robustness. Next, in Setup \hyperref[case:VI]{VI} change the loss function and output representation of the Transformer match decoder $D_{\theta}$ to regression-by-classification, and next in Setup \hyperref[case:VII]{VII} use the robust regression loss. Both these changes further significantly improve performance. This setup constitutes \ours. When we change back to the original ConvNet match decoder in~Setup \hyperref[case:VIII]{VIII} from this final setup, we find that the performance significantly drops, showing the importance of the proposed Transformer match decoder.
\begin{table}
\small
    \centering
    \caption{\textbf{SotA comparison on IMC2022~\citep{image-matching-challenge-2022}.} Measured in mAA (higher is better).}    \label{tab:imc2022}
    \begin{tabular}{ll}
    \toprule
     Method $\downarrow$\quad\quad\quad mAA $\rightarrow$&$@10~\uparrow$\\
     \midrule
SiLK~\citep{gleize2023silk} & 68.6 \\
\midrule
SP~\citep{detone2018superpoint}+SuperGlue~\citep{sarlin2020superglue} &72.4 \\
\midrule
LoFTR~\citep{sun2021loftr}~\tiny{CVPR'21} & 78.3 \\
MatchFormer~\citep{wang2022matchformer}~\tiny{ACCV'22} &78.3 \\
QuadTree~\citep{tang2022quadtree}~\tiny{ICLR'22} &81.7 \\
ASpanFormer~\citep{chen2022aspanformer}~\tiny{ECCV'22} &83.8    \\
DKM~\citep{edstedt2023dkm}~\tiny{CVPR'23} & 83.1\\

\textbf{RoMa} & \textbf{88.0}\\
\bottomrule
    \end{tabular}
\end{table}
\begin{table}
\small
\centering
\caption{\textbf{SotA comparison on WxBS~\citep{Mishkin2015WXBS}.} Measured in mAA at 10px (higher is better). } 

\begin{tabular}{ l  l  }
  \toprule
        Method \quad\quad\quad mAA$@$ $\rightarrow$& $10\text{px}\uparrow$ \\
 \midrule
DISK~\citep{tyszkiewicz2020disk}~\tiny{NeurIps'20}& 35.5\\
 \midrule
DISK + LightGlue~\citep{tyszkiewicz2020disk,lindenberger2023lightglue} ~\tiny{ICCV'23}&41.7\\
SuperPoint +SuperGlue~\citep{detone2018superpoint,sarlin2020superglue}~\tiny{CVPR'20}&31.4\\
 \midrule
LoFTR~\citep{sun2021loftr}~\tiny{CVPR'21} & 55.4\\
DKM~\citep{edstedt2023dkm}~\tiny{CVPR'23} & 58.9\\
\textbf{RoMa}& \textbf{80.1}\\
  \bottomrule
\end{tabular}
\label{tab:WxBS}
\end{table}

\begin{table}
\small
    \centering
    \caption{\textbf{SotA comparison on MegaDepth-1500~\citep{li2018megadepth,sun2021loftr}}. Measured in AUC (higher is better).}
    \begin{tabular}{l lll }
    \toprule
     Method $\downarrow$\quad\quad\quad AUC$@$ $\rightarrow$
     &$5^{\circ}$ $\uparrow$&$10^{\circ}$ $\uparrow$&$20^{\circ}$ $\uparrow$\\
     \midrule         LightGlue~\citep{lindenberger2023lightglue}~\tiny{ICCV'23} & 51.0 & 68.1 & 80.7\\
         \midrule
         LoFTR~\citep{sun2021loftr}~\tiny{CVPR'21}& 52.8 & 69.2 & 81.2 \\
        PDC-Net+~\citep{truong2023pdc}~\tiny{TPAMI'23} & 51.5 & 67.2 & 78.5 \\
         ASpanFormer~\citep{chen2022aspanformer}~\tiny{ECCV'22} & 55.3  & 71.5 & 83.1\\
        ASTR~\citep{yuandchang2023astr}~\tiny{CVPR'23}  & 58.4 & 73.1 & 83.8\\
         DKM~\citep{edstedt2023dkm}~\tiny{CVPR'23}  &  60.4 & 74.9 & 85.1 \\
         PMatch~\citep{zhu2023pmatch}~\tiny{CVPR'23} & 61.4 & 75.7 & 85.7 \\
        CasMTR~\citep{cao2023casmtr}~\tiny{ICCV'23} & 59.1 & 74.3 & 84.8\\
        
         \textbf{\ours}  &  \textbf{62.6} & \textbf{76.7} & \textbf{86.3}\\
    \bottomrule

    \end{tabular}

    \label{tab:megadepth}
\end{table}

\begin{table}
\small
    \centering
    \caption{\textbf{SotA comparison on ScanNet-1500~\citep{dai2017scannet,sarlin2020superglue}}. Measured in AUC (higher is better).}
    \begin{tabular}{l lll lll}
    \toprule
     Method $\downarrow$\quad\quad\quad AUC$@$ $\rightarrow$
     &$5^{\circ}$ $\uparrow$&$10^{\circ}$ $\uparrow$&$20^{\circ}$ $\uparrow$\\
     \midrule
         SuperGlue~\citep{sarlin2020superglue}~\tiny{CVPR'19} & 16.2&	33.8&	51.8\\
        
         \midrule
         LoFTR~\citep{sun2021loftr}~\tiny{CVPR'21}& 22.1 &	40.8&	57.6\\
        PDC-Net+~\citep{truong2023pdc}~\tiny{TPAMI'23} & 20.3 & 39.4 & 57.1\\        
         ASpanFormer~\citep{chen2022aspanformer}~\tiny{ECCV'22} 
         & 25.6 & 46.0 & 63.3\\
        PATS~\citep{ni2023pats}~\tiny{CVPR'23} 
        &  26.0 & 46.9 & 64.3\\
         DKM~\citep{edstedt2023dkm}~\tiny{CVPR'23}  &  29.4 & 50.7 & 68.3\\
         PMatch~\citep{zhu2023pmatch}~\tiny{CVPR'23} & 29.4 & 50.1 & 67.4\\
        CasMTR~\citep{cao2023casmtr}~\tiny{ICCV'23} & 27.1 & 47.0 & 64.4\\
        
         \textbf{\ours}  &   \textbf{31.8} & \textbf{53.4} & \textbf{70.9}\\
    \bottomrule

    \end{tabular}

    \label{tab:scannet}
\end{table}

 \subsection{Training Setup}
We use the training setup as in DKM~\citep{edstedt2023dkm}.
 Following DKM, we use a canonical learning rate (for {\tt batchsize = 8}) of $10^{-4}$ for the decoder, and $5\cdot 10^{-6}$ for the encoder(s). We use the same training split as in DKM, which consists of randomly sampled pairs from the MegaDepth and ScanNet sets excluding the scenes used for testing. The supervised warps are derived from dense depth maps from multi-view-stereo (MVS) of SfM reconstructions in the case of MegaDepth, and from RGB-D for ScanNet. Following previous work~\citep{sun2021loftr,chen2022aspanformer,edstedt2023dkm}, use a model trained on the ScanNet training set when evaluating on ScanNet-1500. All other evaluation is done on a model trained only on MegaDepth.

 As in DKM we train both the coarse matching and refinement networks jointly. Note that since we detach gradients between the coarse matching and refinement, the network could in principle also be trained in two stages. For results used in the ablation, we used a resolution of $448\times 448$, and for the final method we trained on a resolution of $560\times 560$.

\subsection{Two-View Geometry}
We evaluate on a diverse set of two-view geometry benchmarks. We follow DKM~\citep{edstedt2023dkm} and sample correspondences using a balanced sampling approach, producing \num{10000} matches, which are then used for estimation. We consistently improve compared to prior work across the board, in particular achieving a relative error reduction on the competitive IMC2022~\citep{image-matching-challenge-2022} benchmark by 26\%, and a gain of 36\% in performance on the exceptionally difficult WxBS~\citep{Mishkin2015WXBS} benchmark.  

\parsection{Image Matching Challenge 2022:}
We submit to the 2022 version of the image matching challenge~\citep{image-matching-challenge-2022}, which consists of a hidden test-set of Google street-view images with the task to estimate the fundamental matrix between them. We present results in Table~\ref{tab:imc2022}. \ours~attains significant improvements compared to previous approaches, with a relative error reduction of 26\% compared to the previous best approach. 

\parsection{WxBS Benchmark:}
We evaluate \ours~on the extremely difficult WxBS benchmark~\citep{Mishkin2015WXBS}, version 1.1 with updated ground truth and evaluation protocol\footnote{\url{https://ducha-aiki.github.io/wide-baseline-stereo-blog/2021/07/30/Reviving-WxBS-benchmark}}. The metric is mean average precision on ground truth correspondences consistent with the estimated fundamental matrix at a 10 pixel threshold. All methods use MAGSAC++~\citep{magsacpp2019} as implemented in OpenCV. Results are presented in Table~\ref{tab:WxBS}. Here we achieve an outstanding improvement of 36\% compared to the state-of-the-art. We attribute these major gains to the superior robustness of \ours~compared to previous approaches. We qualitatively present examples of this in the supplementary.

\parsection{MegaDepth-1500 Pose Estimation:}
We use the MegaDepth-1500 test set~\citep{sun2021loftr} which consists of 1500 pairs from scene 0015 (St.\ Peter's Basilica) and 0022 (Brandenburger Tor). We follow the protocol in~\citep{sun2021loftr,chen2022aspanformer} and use a RANSAC threshold of 0.5. Results are presented in 
Table~\ref{tab:megadepth}.

\parsection{ScanNet-1500 Pose Estimation:} ScanNet~\citep{dai2017scannet} is a large scale indoor dataset, composed of challenging sequences with low texture regions and large changes in perspective. We follow the evaluation in SuperGlue~\citep{sarlin2020superglue}. Results are presented in 
Table~\ref{tab:scannet}. We achieve state-of-the-art results, achieving the first AUC$@$20$^\circ$ scores over 70.

\parsection{MegaDepth-8-Scenes:}
We evaluate \ours~on the Megadepth-8-Scenes benchmark~\citep{li2018megadepth,edstedt2023dkm}. We present results in Table~\ref{tab:megadepth-8-scenes}. Here too we outperform previous approaches.
\begin{table}
\small
    \centering
    \caption{ \textbf{SotA comparison on Megadepth-8-Scenes~\citep{edstedt2023dkm}.} Measured in AUC (higher is better).}    \label{tab:megadepth-8-scenes}
    \begin{tabular}{llll}
    \toprule
     Method $\downarrow$\quad\quad\quad AUC $\rightarrow$& $@5^{\circ}$&$@10^{\circ}$&$@20^{\circ}$\\
     \midrule
         PDCNet+~\citep{truong2023pdc}~\tiny{TPAMI'23} & 51.8 & 66.6 & 77.2 \\
         ASpanFormer~\citep{chen2022aspanformer}~\tiny{ECCV'22} & 57.2  & 72.1 & 82.9\\
         DKM~\citep{edstedt2023dkm}~\tiny{CVPR'23}  &  60.5 & 74.5 & 84.2\\
\textbf{RoMa}  &  \textbf{62.2} & \textbf{75.9} & \textbf{85.3}\\
    \bottomrule
    \end{tabular}
\end{table}


\subsection{Visual Localization}
We evaluate \ours~on the InLoc~\citep{taira2018inloc} Visual Localization benchmark, using the HLoc~\citep{sarlin2019coarse} pipeline. We follow the approach in DKM~\citep{edstedt2023dkm} to sample correspondences. Results are presented in Table~\ref{tab:inloc}. We show large improvements compared to all previous approaches, setting a new state-of-the-art. 
\begin{table}
\small
    \caption{\textbf{SotA comparison on InLoc~\citep{taira2018inloc}.} We report the percentage of query images localized within 0.25/0.5/1.0 meters and 2/5/10$^{\circ}$ of the ground-truth pose (higher is better).}

    \centering
    \begin{tabular}{l c c}
    \toprule
Method & DUC1 & DUC2\\
& \multicolumn{2}{c}{(0.25m,2$^{\circ}$)/(0.5m,5$^{\circ}$)/(1.0m,10$^{\circ}$)}  \\
\midrule
PATS &55.6 / 71.2 / 81.0 &	58.8 
 / 80.9 / 85.5\\
DKM & 51.5 / 75.3 / 86.9 &	63.4 / 82.4 / \textbf{87.8} \\
CasMTR & 53.5 / 76.8 / 85.4 & 51.9 / 70.2 / 83.2\\
\textbf{RoMa}  & \textbf{60.6} / \textbf{79.3} / \textbf{89.9} &	\textbf{66.4} / \textbf{83.2} / \textbf{87.8} \\
\bottomrule
    \end{tabular}
    \label{tab:inloc}
\end{table}



\subsection{Runtime Comparison}
We compare the runtime of RoMa and the baseline DKM at a resolution of $560\times560$ at a batch size of 8 on an RTX6000 GPU. We observe a modest 7$\%$ increase in runtime from 186.3 $\to$ 198.8 ms per pair.

\section{Conclusion}
We have presented \ours, a robust dense feature matcher. Our model leverages frozen pretrained coarse features from the foundation model DINOv2 together with specialized ConvNet fine features, creating a precisely localizable and robust feature pyramid. We further improved performance with our proposed tailored transformer match decoder, which predicts anchor probabilities instead of regressing coordinates. Finally, we proposed an improved loss formulation through regression-by-classification with subsequent robust regression. Our comprehensive experiments show that \ours~achieves major gains across the board, setting a new state-of-the-art. In particular, our biggest gains (36\% increase on WxBS~\citep{Mishkin2015WXBS}) are achieved on the most difficult benchmarks, highlighting the robustness of our approach. Code is provided at \href{https://github.com/Parskatt/RoMa}{github.com/Parskatt/RoMa}.

\parsection{Limitations and Future Work:}
\begin{enumerate}[topsep=0.5ex,itemsep=0ex,label=(\alph*)]
    \item Our approach relies on supervised correspondences, which limits the amount of usable data. We remedied this by using pretrained frozen foundation model features, which improves generalization.
    \item We train on the task of dense feature matching which is an indirect way of optimizing for the downstream tasks of two-view geometry, localization, or 3D reconstruction. Directly training on the downstream tasks could improve performance.
\end{enumerate}

\newpage
{
    \small
    \bibliographystyle{ieeenat_fullname}
    \bibliography{main}
}

\clearpage
\setcounter{page}{1}
\maketitlesupplementary

\appendix
\nocite{balntas2017hpatches}
\begin{figure}[ht]
    \centering
    \includegraphics[width=.45\linewidth]{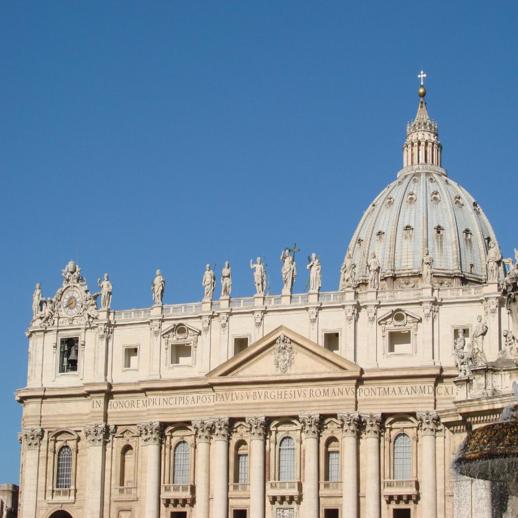}
    \includegraphics[width=.45\linewidth]{figures/linearprobe/im_B.jpg}
    \includegraphics[width=.45\linewidth]{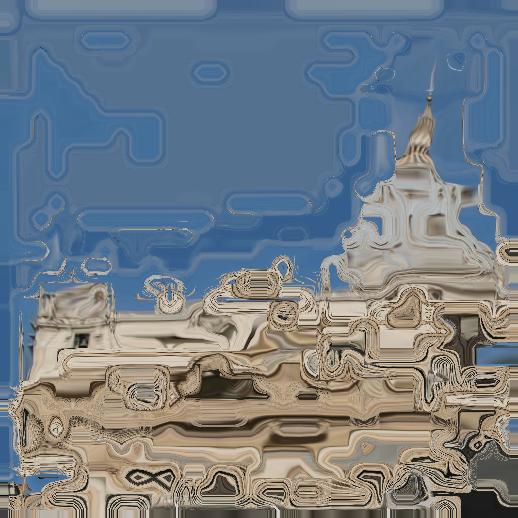}
    \includegraphics[width=.45\linewidth]{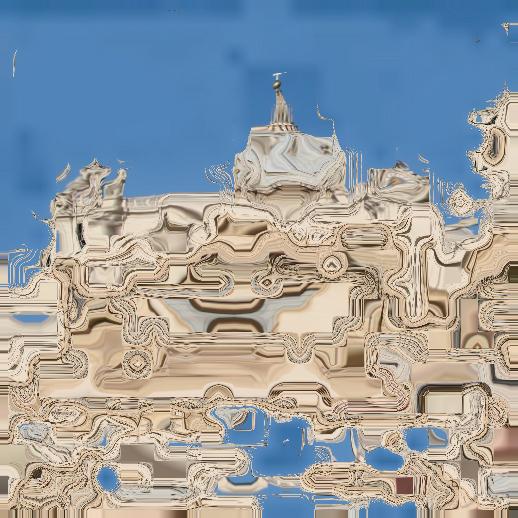}
    \includegraphics[width=.45\linewidth]{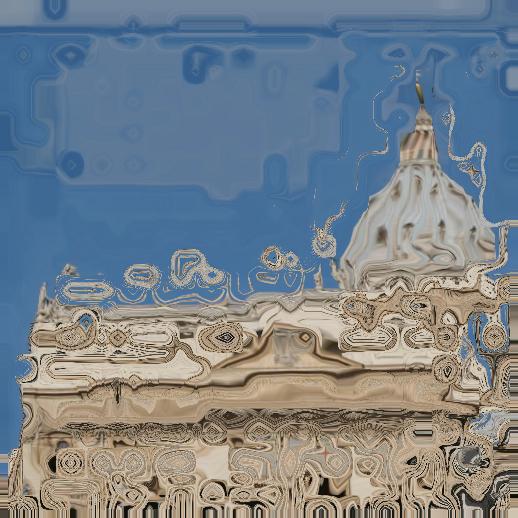}
    \includegraphics[width=.45\linewidth]{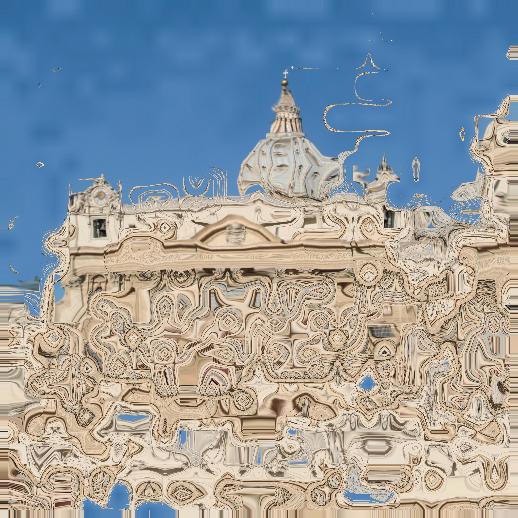}
    \includegraphics[width=.45\linewidth]{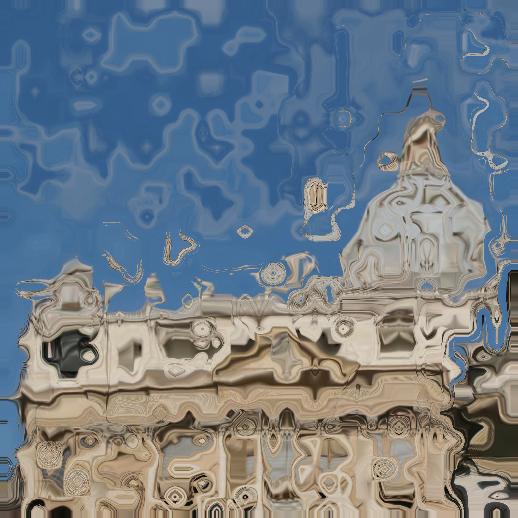}
    \includegraphics[width=.45\linewidth]{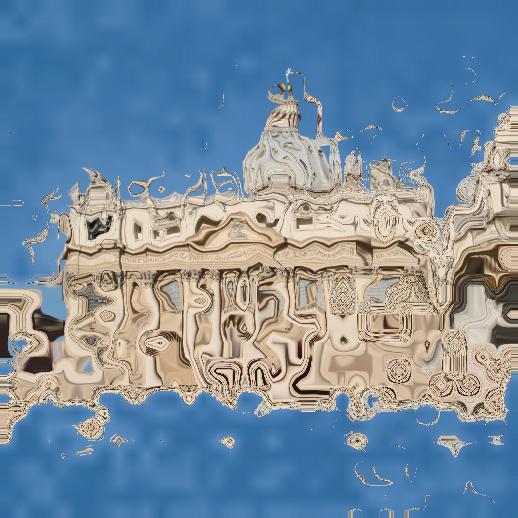}
    \includegraphics[width=.9\linewidth]{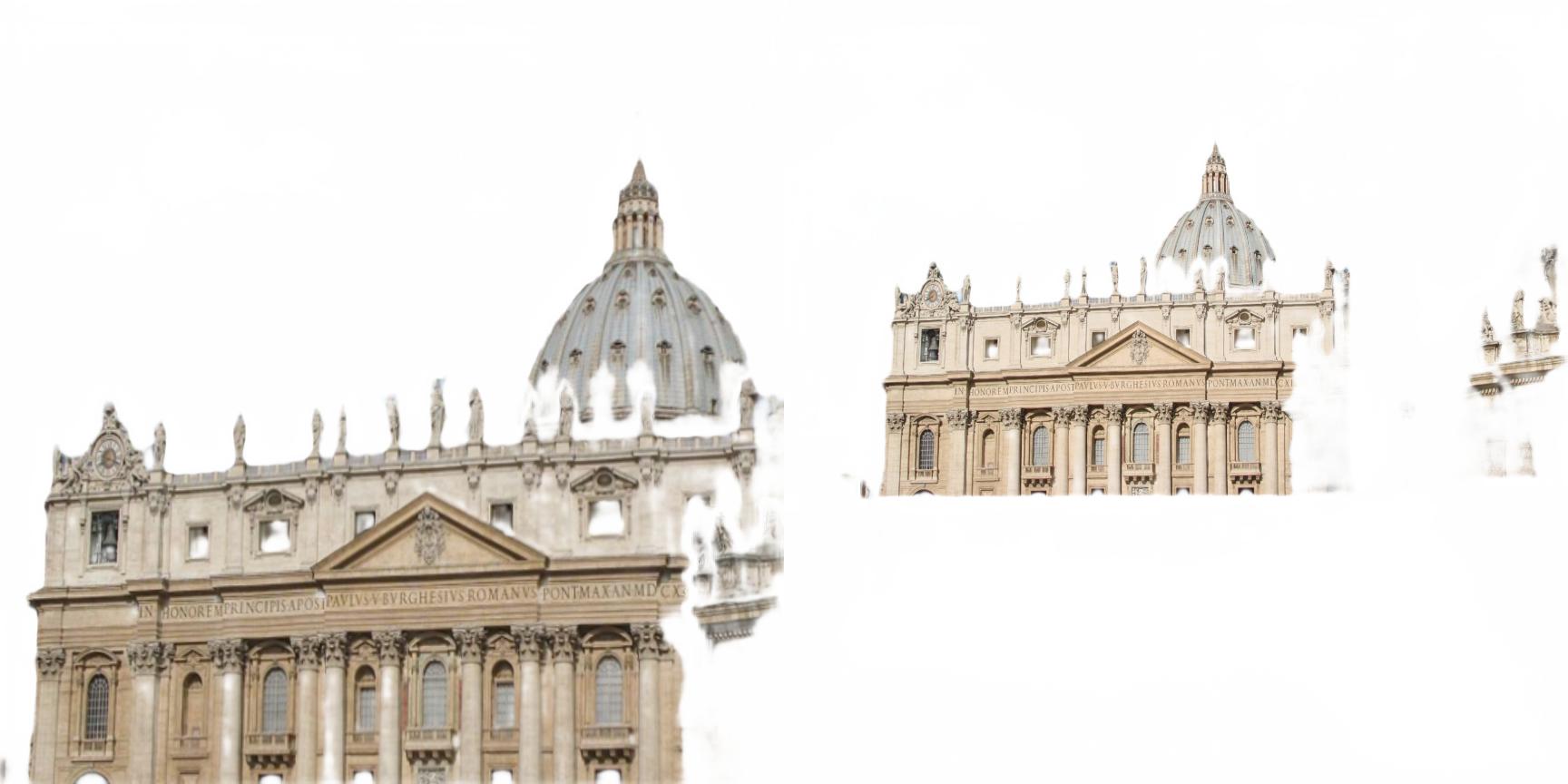}
    \caption{\textbf{Evaluation of frozen features.} From top to bottom: Image pair, VGG19 matches, RN50 matches, DINOv2 matches, RoMa matches. DINOv2 is significantly more robust than the VGG19 and RN50. Quantitative results are presented in Table~\ref{tab:linprobe}.}
    \label{fig:frozen-features-supp}
\end{figure}

In this supplementary material, we provide further details and qualitative examples that could not fit into the main text of the paper.
\section{Further Details on Frozen Feature Evaluation}
We use an exponential cosine kernel as in DKM~\citep{edstedt2023dkm} with an inverse temperature of 10. We train using the same training split as in our main experiments, using the same learning rates (note that we only train a single linear layer, as the backbone is frozen). We use the regression-by-classification loss that we proposed in Section~\ref{sec:loss}. We present a qualitative example of the estimated warps from the frozen features in Figure~\ref{fig:frozen-features-supp}.

\section{Further Architectural Details}

\parsection{Encoders:}
We extract fine features of stride $\{1,2,4,8\}$ by taking the outputs of the layer before each $2\times 2$ maxpool. These have dimension $\{64,128,256,512\}$ respectively. We project these with a linear layer followed by batchnorm to dimension $\{9,64,256,512\}$.

We use the patch features from DINOv2~\citep{oquab2023dinov2} and do not use the \texttt{cls} token. We use the ViT-L-14 model, with patch size 14 and dimension $1024$. We linearly project these features (with batchnorm) to dimension $512$.

\parsection{Global Matcher:}
We use a Gaussian Process~\citep{rasmussen2003gaussian} match encoder as in DKM~\citep{edstedt2023dkm}. We use an exponential cosine kernel~\citep{edstedt2023dkm}, with inverse temperature 10. As in DKM, the GP predicts a posterior over embedded coordinates in the other image. We use an embedding space of dimension $512$.

For details on $D_{\theta}$ we refer to Section~\ref{sec:transformer-decoder}.

\parsection{Refiners:} Following \citet{edstedt2023dkm} we use 5 refiners at strides $\{1,2,4,8,14\}$. They each consist of 8 convolutional blocks. The internal dimension is set to $\{24,144,569,1137,1377\}$. The input to the refiners are the stacked feature maps, local correlation around the previous warp of size $\{0,0,5,7,15\}$, as well as a linear encoding of the previous warp. The output is a $B \times H_s\times W_s \times (2+1)$ tensor, containing the warp and an logit offset to the certainty.

\section{Qualitative Comparison on WxBS}
\begin{figure*}
    \centering
    \includegraphics[width=.8\linewidth]{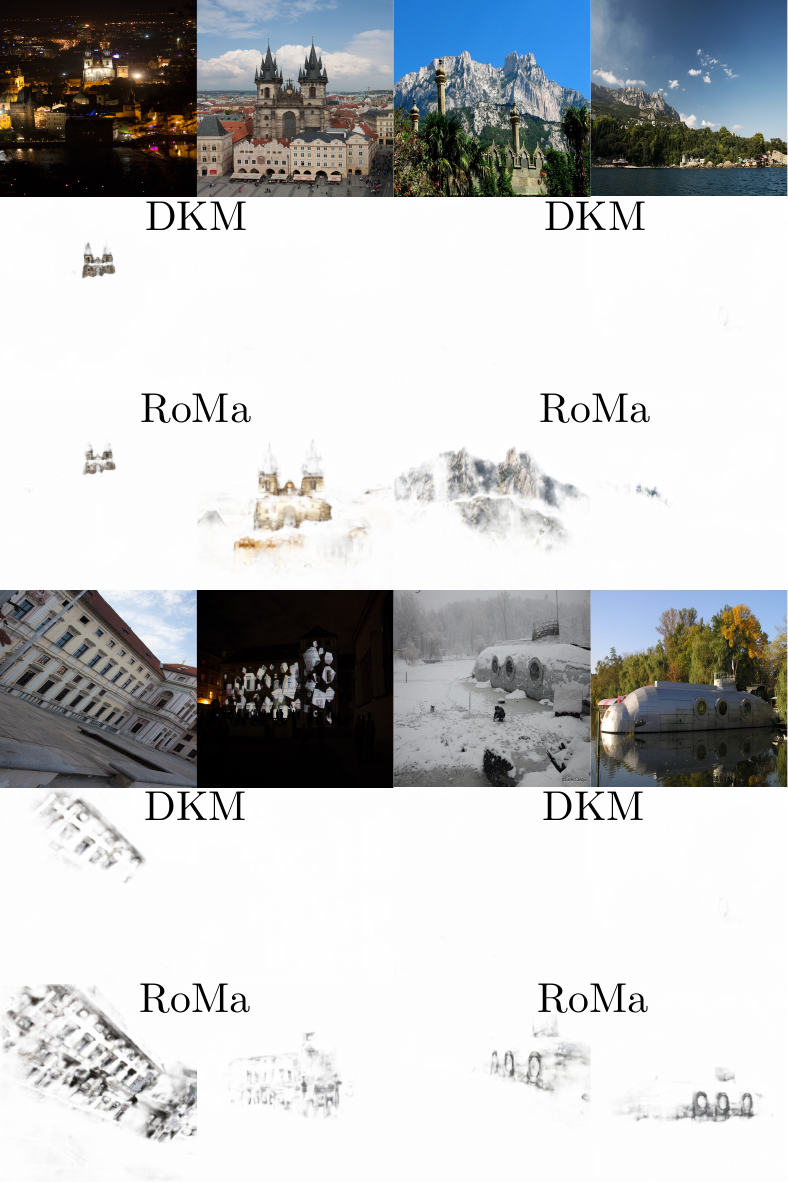}
    \caption{\textbf{Qualitative comparison.} RoMa is significantly more robust to extreme changes in viewpoint and illumination than DKM.}
    \label{fig:qualitative-supp}
\end{figure*}

We qualitatively compare estimated matches from RoMa and DKM on the WxBS benchmark in Figure~\ref{fig:qualitative-supp}. DKM fails on multiple pairs on this dataset, while RoMa is more robust. In particular, RoMa is able to match even for changes is season (bottom right), extreme illumination (bottom left, top left), and extreme scale and viewpoint (top right).

\section{Further Details on Metrics}
\parsection{Image Matching Challenge 2022:} The mean average accuracy (mAA) metric is computed between the estimated fundamental matrix and the hidden ground truth. The error in terms of rotation in degrees and translation in meters. Given one threshold over each, a pose is classified as accurate if it meets both thresholds. This is done over ten pairs of uniformly spaced thresholds. The mAA is then the average over the threshold and over the images (balanced across the scenes).

\parsection{MegaDepth/ScanNet:} The AUC metric used measures the error of the estimated Essential matrix compared to the ground truth. The error per pair is the maximum of the rotational and translational error. As there is no metric scale available, the translational error is measured in the cosine angle. The recall at a threshold $\tau$ is the percentage of pairs with an error lower than $\tau$. The AUC$@\tau^{\circ}$ is the integral over the recall as a function of the thresholds, up to $\tau$, divided by $\tau$. In practice, this is approximated by the trapezial rule over all errors of the method over the dataset.

\section{Further Details on Theoretical Model}
Here we discuss a simple connection to scale-space theory, that did not fit in the main paper.
Our theoretical model of matchability in Section~\ref{sec:loss} has a straightforward connection to scale-space theory~\citep{witkin1983scale,koenderink1984structure,lindeberg1994scale}. The image scale-space is parameterized by a parameter $s$,
\begin{equation}
    L(x,s) = \int g(x-y;s)I(y)dy,
\end{equation}
 where \begin{equation}
     g(x;s)=\frac{1}{2\pi s^2}\exp\bigg(-\frac{1}{2s^2}\lVert x\rVert^2\bigg)
 \end{equation} is a Gaussian kernel. Applying this kernel jointly on the matching distribution yields the diffusion process in the paper.

\section{Further Details on Match Sampling}
Dense feature matching methods produce a dense warp and certainty. However, most robust relative pose estimators (used in the downstream two-view pose estimation evaluation) assume a sparse set of correspondences. While one could in principle use all correspondences from the warp, this is prohibitively expensive in practice. We instead follow the approach of DKM~\citep{edstedt2023dkm} and use a balanced sampling approach to produce a sparse set of matches. The balanced sampling approach uses a KDE estimate of the match distribution $p_{\theta}(x^{\A}, x^{\B})$ to rebalance the distribution of the samples, by reweighting the samples with the reciprocal of the KDE. This increases the number of matches in less certain regions, which~\citet{edstedt2023dkm} demonstrated improves performance.

\newpage

\end{document}